\documentclass{bmvc2k}
\usepackage{wrapfig}
\usepackage{bm}
\usepackage{pifont}
\usepackage{multirow}
\usepackage{graphicx,subfigure}
\usepackage{pifont}% http://ctan.org/pkg/pifont
\usepackage{algorithm}
\usepackage{algpseudocode}
\usepackage{booktabs}       % professional-quality tables
\usepackage{color}
\usepackage{colortbl}
\usepackage[outercaption]{sidecap}
\usepackage{caption}

\definecolor{Gray}{gray}{0.90}
\definecolor{white}{rgb}{1.0, 1.0, 1.0}

\definecolor{LightCyan}{RGB}{247, 223, 231}
\newcolumntype{a}{>{\columncolor{LightCyan}}c}
\definecolor{Gray}{gray}{0.90}
\title{How to Train Vision Transformer on Small-scale Datasets?}

\addauthor{Hanan Gani}{hanan.ghani@mbzuai.ac.ae}{1}
\addauthor{Muzammal Naseer}{muzammal.naseer@mbzuai.ac.ae}{1}
\addauthor{Mohammad Yaqub}{mohammad.yaqub@mbzuai.ac.ae}{1}

\addinstitution{
 Mohamed Bin Zayed University of Artificial Intelligence,\\
 Abu Dhabi, UAE\\
}

\runninghead{Gani et. al.}{How to Train ViT on Small-scale Datasets?}

\def\eg{\emph{e.g}\bmvaOneDot}

\usepackage{tikz}
\usetikzlibrary{shapes.misc,shadows}

\DeclareRobustCommand{\tcblack}[1]{
\begin{tikzpicture}[baseline=(char.base)]
\node(char)[
  draw,fill=black,
  shape=rounded rectangle,
  text height=5pt,
  drop shadow={opacity=.5,shadow xshift=0pt,shadow yshift=-1pt},
]
  {\color{white}{\normalfont #1}};
\end{tikzpicture}
}

\begin{document}

\maketitle
\begin{abstract}
  Vision Transformer (ViT), a radically different architecture than convolutional neural networks offers multiple advantages including design simplicity, robustness and state-of-the-art performance on many vision tasks. However, in contrast to convolutional neural networks, Vision Transformer lacks inherent inductive biases. Therefore, successful training of such models is mainly attributed to pre-training on large-scale datasets such as ImageNet with 1.2M  or JFT  with 300M images. This hinders the direct adaption of Vision Transformer for small-scale datasets. In this work, we show that self-supervised inductive biases can be learned directly from small-scale datasets and serve as an effective weight initialization scheme for fine-tuning. This allows to train these models without large-scale pre-training, changes to model architecture or loss functions. We present thorough experiments to successfully train monolithic and non-monolithic Vision Transformers on five small datasets including CIFAR10/100, CINIC10, SVHN, Tiny-ImageNet and two fine-grained datasets: Aircraft and Cars. Our approach consistently improves the performance of Vision Transformers while retaining their properties such as attention to salient regions and higher robustness. Our codes and pre-trained models are available at: \url{https://github.com/hananshafi/vits-for-small-scale-datasets}.

\end{abstract}

%------------------------------------------------------------------------- 
\section{Introduction}
\label{sec:intro}
Since their inception, Vision Transformers (ViTs) \cite{khan2021transformers, alexey2020vit, park2022how} have emerged as an effective alternative to traditional Convolutional Nerual Networks (CNNs) \cite{kaiming2015resnet, huang2016densenet, Mingxing2019EfficientNet, NIPS2012_c399862d, szegedy2014deeper,wang2019deephigh}. The architecture of Vision Transformer is inspired by the prominent Transformer encoder \cite{transformer2017nlp,devlin2018bert} used in natural language processing (NLP) tasks, which process data in the form of sequence of vectors or tokens.
Similar to the word tokens in NLP Transformer,  ViT typically splits the image into a grid of non-overlapping patches before passing them to a linear projection layer to adjust the token dimensionality. These tokens are then processed by a series of feed-forward and multi-headed self-attention layers. Due to their ability to capture global structure through self-attention \cite{naseer2021intriguing}, ViTs have found extensive applications in many tasks such as classification \cite{alexey2020vit, touvron2020deit, yuan2021t2t,liu2021swin,wu2021cvt,yuan2021cvdesign, li2021localvit, xu2021coscale,touvron2021cait} object detection \cite{carion2020detr,kamath2021mdetr,zhu2020defordeter,dai2020updetr}, segmentation \cite{strudel2021segmenter,rabftl2021densepred}, restoration \cite{liang2021restoration}, and 3D vision \cite{CHEN2021MVT,zhao2022point}.

\begin{figure}[!t]
 \small \centering
 \begin{minipage}{0.45\textwidth}
 %\fbox{\rule[-.5cm]{0cm}{3cm} \rule[-.5cm]{3cm}{0cm}}
  \includegraphics[width=\linewidth]{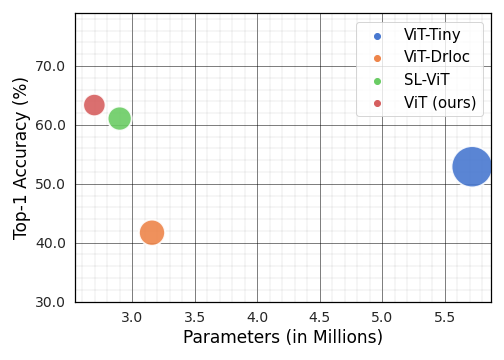}
 \end{minipage}
 \hfill
  \begin{minipage}{0.54\textwidth}
    \caption{\small Our approach is simple in nature and yet outperforms \cite{touvron2020deit, lee2021vitsmall, liu2021nipsvitsmall} by notable margins both in terms of  trainable parameters and generalization (top-1 accuracy) on Tiny-ImageNet \cite{Le2015TinyImagenet}. This shows that inductive biases learned by our self-supervised approach serves an effective weight initialization to ease ViT optimization during supervised training. Our approach remains beneficial to different ViT designs as we thoroughly explored on multiple small-scale datasets in Sec. \ref{subsec: Results}
  }
\label{fig: into fig to showcase performance}
  \end{minipage}
\vspace{-2em}
\end{figure}

Despite their advantages, ViTs fail to match the performance of CNNs when trained from scratch on small-scale datasets. This is primarily due to the lack of locality, inductive biases and hierarchical structure of the representations which is commonly observed in the CNN architectures \cite{liu2021swin, wu2021cvt,yuan2021cvdesign}. As a result, ViTs require large-scale pre-training to learn such properties from the data \cite{alexey2020vit} for better transfer learning to downstream tasks. Typically, ViTs are trained with a private JFT-300M \cite{SUN2017JFT} dataset with 303 million weakly supervised images or publicly available ImageNet-1k/22k datasets \cite{deng2009imagenet}. However, the absence of such large-scale pre-training hampers the performance of ViTs on small-scale datasets \cite{lee2021vitsmall, liu2021nipsvitsmall}.

To ease the optimization difficulties during ViT training, different architectural designs are proposed to induce the necessary inductive biases for the Vision Transformer \cite{graham2021levit,liu2021swin,wu2021cvt,yuan2021cvdesign,li2021localvit}. These hybrid networks still remain sub-optimal for small datasets (Sec. \ref{subsec: Results}) and either require further modifications to the loss functions \cite{liu2021nipsvitsmall} or network architecture \cite{lee2021vitsmall}. Even with careful design choices, these methods remain sensitive to the type of data distribution \eg, \cite{liu2021nipsvitsmall}'s performance degrades on Tiny-ImageNet, a complex data distribution as compared to other small datasets such CIFAR (Sec. \ref{subsec: Ablative Analysis}).

 To alleviate these problems, we propose an effective two-stage framework to train ViTs on small-scale low resolution datasets from \emph{scratch}. \textbf{Low-resolution View Prediction as Weight Initialization Scheme:} We observe that ViTs are sensitive to weight initialization and converge to vastly different solutions depending on the network initialization (Sec. \ref{sec: Vision Transformer Training on Small Datasets}).
 %as compared to CNNs (Sec. \ref{sec: Vision Transformer Training on Small Datasets}) due to lack of inductive biases. 
 Large-scale pre-training captures inductive biases from the data \cite{alexey2020vit} and allows successful transfer learning on small datasets. In the absence of huge datasets, however,  we hypothesize that ViTs can benefit from the inductive biases directly learned on the target small dataset such as CIFAR10 or CIFAR100. To this end, we introduce self-supervised weight learning scheme based on feature prediction of our \emph{low-resolution} global and local views via self-distillation \cite{caron2021dino}. \textbf{Self-supervised to Supervised Learning for small-scale Datasets:} In the second stage, we fine-tune the same ViT network on the same target dataset using simply cross-entropy loss. Our approach therefore is agnostic to ViT architectures, independent to changes in loss functions, and provides significant gains in comparison to different weight initialization schemes \cite{pmlr-v9-glorot10a, kaiming2015heini, NEURIPS2019_9015_pytorch, renzato2021gradinit} and existing works \cite{lee2021vitsmall, liu2021nipsvitsmall} (Fig. \ref{fig: into fig to showcase performance}). We demonstrate the effectiveness of our method on five small datasets across different monolithic and non-monolith Vision Transformers (Sec. \ref{subsec: Results}). Our contributions are as follows:

\begin{enumerate}
 \item We propose a  self-supervised weight learning scheme from low-resolution views created on small datasets. This serves as an effective weights initialization to successfully train ViTs from scratch, thus eliminating the need for large-scale pre-training.

\item Our proposed self-supervised inductive biases  improve the performance of ViTs on small datasets without modifying the network architecture or loss functions.

\item We show that our training approach scales well with the input resolution. For instance, when trained on high-resolution samples, our method improves by 8\% (CIFAR10)  and 7\% (CIFAR100) w.r.t the state-of-the-art baseline \cite{liu2021nipsvitsmall} for training ViTs on small datasets.  Furthermore, we validate the efficacy of our method by observing  its robustness against natural corruptions, and attention to salient regions in the input sample. 
\end{enumerate}

\section{Related Work}
In this section, we discuss the related existing works in the application of ViTs for small datasets, self-supervised learning and weight initialization. There have been attempts to train ViTs on ImageNet from scratch \cite{touvron2020deit,graham2021levit,yuan2021t2t,wu2021cvt, touvron2021cait,liu2021swin}. \cite{touvron2020deit} improves the performance of ViT through data augmentations, regularization, and knowledge distillation.
\cite{yuan2021t2t} introduces a new image tokenization strategy by recursively aggregating the neighboring tokens in order to model the locality into the network. \cite{liu2021swin} introduces a hierarchical vision transformer which processes the input at various scales and limits the self-attention to non-overlapping patches by the use of shifted windows. \cite{wu2021cvt} replaces the projection and multilayer perceptron layers with convolution layers in order to introduce the shift, scale, and distortion invariance.  Recently, there have been few attempts to train ViTs on small datasets \cite{lee2021vitsmall, liu2021nipsvitsmall}.

\noindent\textbf{Vision Transformers for Small Datasets:}  \cite{lee2021vitsmall} applies a series of augmentations \cite{dan2019augmix, yun2019cutmix, zhong2017erasing, cubuk2019randaug, szegedy2015labelsmooth} on the input data and introduces shifted patch tokenization (SPT) and locality self-attention (LSA), which enable ViT to learn from scratch even on small datasets.
\cite{liu2021nipsvitsmall} trains a ViT with an additional proxy task of learning the spatial location of the encoded image tokens in order to learn the phenomena of locality.
Different from these approaches, we show that without any modification to the internal layers or addition of new loss function, our approach learns better generalizable features from the existing small target datasets.

\noindent\textbf{Self-supervised learning:} In recent years, several self-supervised techniques have been proposed to pre-train ViTs
\cite{caron2021dino, atito2021sit, zhou2021ibot,li2021esvit, kaiming2021maskedae,xie2021simmmim}. In \cite{caron2021dino}, the pretext task is to match the local and global features by minimizing the cross entropy loss. In \cite{kaiming2021maskedae, xie2021simmmim}, the input patches are masked and the network is tasked to predict the masked pixels. \cite{zhou2021ibot} pretrains the network with two pretext tasks based on local-global feature matching and masked encoding. All these methods have shown impressive results on Imagenet linear evaluation and have been applied to numerous downstream tasks. Such pre-training strategies are computationally expensive and are designed for large sized datasets at higher resolution. Instead, we apply self-supervision for low-resolution small dataset to observe decent improvements.

\noindent\textbf{Weight Initialization.} ImageNet pre-trained weights have been the default choice for network initialization in most computer vision tasks. However, given the amount of training time and computational resources required for such training, some past works \cite{pmlr-v119-huang20f, renzato2021gradinit} have proposed methods to efficiently initialize the model weights.  \cite{pmlr-v119-huang20f} introduces a weight initialization scheme that eliminates the problem of learning rate warmup in NLP transformers, enabling deep transformer models to train without difficulty. \cite{renzato2021gradinit} presents a model agnostic initialization scheme which adjusts the norm of each network layer by introducing a multiplier variable in front of each parameter block. Apart from these approaches, majority of the models are initialized using the basic weight initialization schemes \cite{pmlr-v9-glorot10a, kaiming2015heini, NEURIPS2019_9015_pytorch}, etc. 
Different from these approaches, we learn the initial weights of the ViT using self-supervised learning  directly from small datasets
without any changes in the architecture or the optimizer.

\begin{figure}[!t]
 \small \centering
 \begin{minipage}{0.7\textwidth}
 % lelf lower right up 
  \includegraphics[width=\linewidth, trim= 0mm 0mm 2.5mm 0mm, clip]{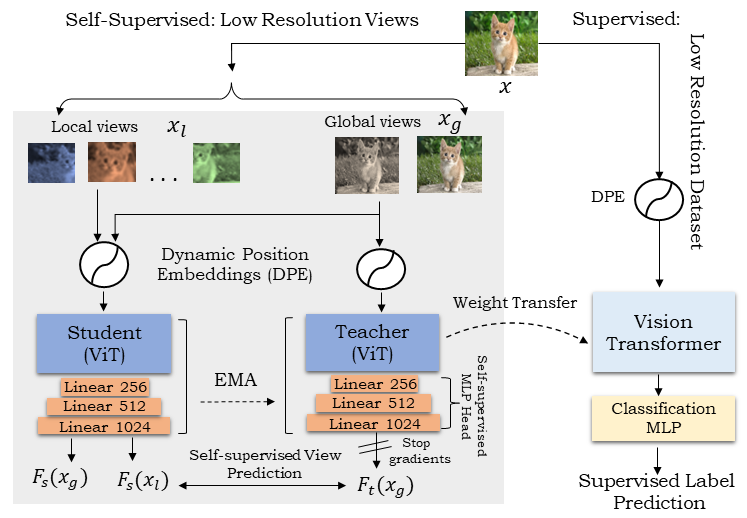}
 \end{minipage}
 \hfill
  \begin{minipage}{0.28\textwidth}
    \caption{\small Our Training framework: At the pre-training stage, we learn self-supervised weights by predicting  \emph{low resolution} local and global views via self-distillation through student and teacher networks \cite{caron2021dino}. At the fine-tuning stage,  self-supervised weights are used to initialize ViT for supervised learning. Our self-supervised inductive biases ease ViT optimization during supervised learning on small-scale datasets.
  }
\label{fig: Our Training framework}
  \end{minipage}
\vspace{-1.2em}
\end{figure}

\begin{figure}[!t]
 \small \centering
 % lelf lower right up 
\includegraphics[width=\linewidth, trim= 0mm 5mm 1.0mm 0mm, clip]{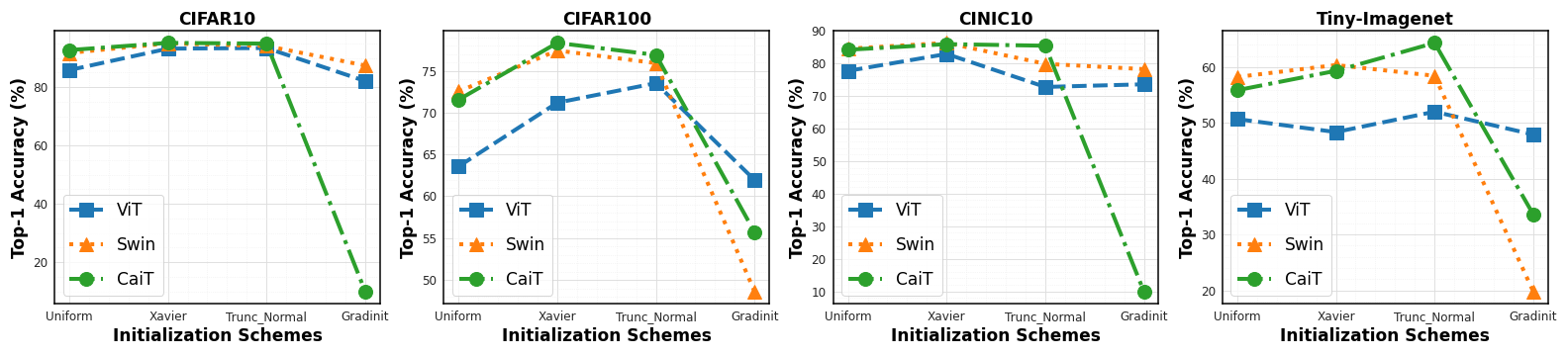}
\vspace{0em} 
\caption{\small We train Vision Transformers  with different weight initialization schemes including `Uniform', `Xavier \cite{pmlr-v9-glorot10a}', `Truncated normal \cite{kaiming2015heini}', and `Gradinit \cite{renzato2021gradinit}'. All models are trained for 100 epochs.  We  follow the default training setting for small-scale datasets as proposed by \cite{lee2021vitsmall} in all our experiments for consistent comparisons. We observe that ViT training can be unstable depending upon weight initializations \eg CaiT \cite{touvron2021cait} performs poorly when initialized with Gradinit. Similarly, the generalization of ViT \cite{touvron2020deit} and Swin \cite{liu2021swin}  varies a lot with different weight initialization methods. 
  }
\label{fig: effect of weight initialization schemes}
\vspace{-1.5em}
\end{figure}

\section{Vision Transformer Training on Small Datasets}
\label{sec: Vision Transformer Training on Small Datasets}
Inherent inductive biases allow to train CNNs on small-scale datasets from scratch. Vision Transformer on the other hand need large-scale pre-training for successful transfer learning \cite{alexey2020vit, touvron2020deit}. Our goal is to eliminate the large-scale data requirement and train ViTs directly on a given small dataset. Neural networks are sensitive to weight initialization schemes \cite{hanin2018start}.  We observe that ViTs' convergence is affected by the weight initialization scheme (Fig. \ref{fig: effect of weight initialization schemes}) during training from scratch. Therefore, we propose to learn the weight initialization from the given data distribution ($\mathcal{Q}$) to inject the necessary inductive biases within ViT architecture. Our training approach (Fig. \ref{fig: Our Training framework}), hence, consists of two stages including  \emph{\textbf{a) Self-supervised View Prediction}} followed by  \emph{\textbf{b) Supervised Label Prediction}} tasks.  Note that both of these tasks are learned on the same data distribution ($\mathcal{Q}$) with the same model backbone $\mathcal{F}$. The only architectural difference between the both learning tasks is the self-supervised and supervised MLP projection (Fig. \ref{fig: Our Training framework}). In this manner, our approach is independent to large-scale pre-training. We describe the ViT encoders designs next in Sec. \ref{subsec: Vision Transformer Encoders} before explaining our proposed training in detail (Sec. \ref{subsec: Self-supervised View Prediction as Weight Initialization Scheme} and \ref{subsec: Self-supervised to Supervised Label Prediction} ).

\subsection{Vision Transformer Encoders}
\label{subsec: Vision Transformer Encoders}
We train different monolithic \cite{touvron2020deit} and non-monolithic (Swin \cite{liu2021swin} and CaiT \cite{touvron2021cait}) ViTs (Table \ref{tab: architectures}). These ViTs are originally designed for higher resolution inputs (224 or 384) with patch sizes of 16 or 32. However, small-scale datasets have low resolution inputs \eg 32 or 64 in the case of CIFAR and Tiny-ImageNet, respectively. Therfore, we reduce the patch size for such low resolution inputs. Specifically, we set a patch size of 8 and 4 for an input of size  64x64 and 32x32, respectively. Similarly, we adopt the original ViT designs for small datasets following \cite{raghu2021vitcnn}. Table \ref{tab: architectures} presents the high level details of these network architectures.  Further ablations with different ViT attributes (\eg depth, and heads) are provided in Appendix \ref{sec:vit-vs-vit-full}.

\begin{table}[!t]
  \begin{minipage}{0.75\textwidth}
   \centering \small
  \setlength{\tabcolsep}{3pt}
  \scalebox{0.8}[0.8]{
  \begin{tabular}{c|cccccc}
  \toprule
  \rowcolor{Gray} 
    \textbf{Attributes $\rightarrow$}   & \textbf{Depth} & \textbf{Patch-size} & \textbf{Token Dimension} & \textbf{Heads} & \textbf{MLP-ratio} & \textbf{Window-size} \\
    \midrule
    ViT \cite{touvron2020deit} & 9 & [4,8] & 192 & 12 & 2 & -\\
    Swin \cite{liu2021swin} & [2,4,6] & [2,4] & 96 & [3,6,12] & 2 & 4\\
    CaiT \cite{touvron2021cait} & 24 & [4,8] & 192&4&2&- \\
  \bottomrule
  \end{tabular}}
  \end{minipage}
  \hfill
  \begin{minipage}{0.22\textwidth}
  \caption{\small Details of ViT encoders used in our proposed training approach (Fig. \ref{fig: Our Training framework}).}
 \label{tab: architectures}
  \end{minipage}
 \vspace{-1em}
\end{table}

\subsection{Self-supervised View Prediction as Weight Initialization Scheme}
\label{subsec: Self-supervised View Prediction as Weight Initialization Scheme}

As mentioned earlier, we learn to initialize wights for low resolution small-scale datasets via self-supervised training. Among many self-supervised learning methods \cite{chen2021mocov3, ranasinghe2022selfsupervised, ting2020simclr}, we adopt view prediction strategy based on \cite{caron2021dino}. Thus, our self-supervised approach does not require memory bank, large batch-size, or negative mining. The self-supervised weights are then used for initialization during fine-tuning stage directly from the low-resolution dataset.
Our view prediction pre-training uses a student ($\mathcal{F}_s$) and teacher ($\mathcal{F}_t$) setup to predict different views of the same input sample from each other and thus follows the learning paradigm of knowledge self-distillation \cite{grill2020byol}. Both student and teacher represent the same network (Fig. \ref{fig: Our Training framework}) but process different views as explained below.

\noindent\textbf{Self-supervised View Generation and Prediction:} Consider a low resolution input $\bm{x}$ sampled from a small data distribution $\mathcal{Q}$. We define the height and width of the low-resolution input $\bm{x}$ by $h$ and $w$, respectively. During pre-training, the input is distorted and augmented to generate global ($\bm{x}_g$) and local ($\bm{x}_l$) views. We use standard augmentations \cite{caron2021dino} which preserve the semantic information of each selected view. These augmentations include color jitter, gray scaling, solarization, random horizontal flip and gaussian blur. \emph{Global views} are generated by randomly selecting regions in the input image covering more than 50\%  of the input portion, while \emph{local views} are generated by randomly selecting regions covering around 20-50\% portion of the input. The global and local views are further resized such that the ratio of area of local to global view is 1:4. As for instance, the global view generated for CIFAR sample is resized to a dimension of 32x32 and the local view is resized to 16x16. 
We generate 2 global and 8 local views in our case.  \emph{Dynamic Position Embeddings (DPE):} The number of input tokens vary based on the view size, so we use Dynamic Position Embeddings (DPE) \cite{caron2021dino, ranasinghe2022selfsupervised} which interpolates for the missing tokens of smaller views with height and width less than the original sample size $h\times w$. Both  student and teacher networks process these multi-sized views and output the corresponding feature representations. \emph{Self-supervised MLP Projection:} The features representation of each view is further processed by a 3-layer MLP of the student and teacher networks. The multi-layer projection performs better than a single layer MLP \cite{caron2021dino}. Thus, each low-resolution view is converted into $1024$ dimensional feature vector. Ablative analysis on the effect of the output size of self-supervised MLP projection head is provided in Sec. \ref{subsec: Ablative Analysis}.

The teacher network processes the global views to generate target features ($\mathcal{F}_{g_{t}}$) while all the  local and global views are forward-passed through the student network to generate predicted features ($\mathcal{F}_{g_{s}}$) and ($\mathcal{F}_{l_{s}}$). These features are normalized \cite{caron2021dino, ranasinghe2022selfsupervised} to obtain $\Tilde{\mathcal{F}_{g_{t}}}, \Tilde{\mathcal{F}_{g_{s}}}$, and $\Tilde{\mathcal{F}_{l_{s}}}$.
We update the student's parameters by minimizing the following objective:

\begin{align}\label{eq: overall_loss}
    \mathcal{L} &=  -\Tilde{\mathcal{F}}_{g_{t}} \cdot \log(\Tilde{\mathcal{F}}_{g_{s}}) + \sum_{i=1}^{n} -\Tilde{\mathcal{F}}_{g_{t}} \cdot \log(\Tilde{\mathcal{F}}^{(i)}_{l_{s}}),
\end{align}
where $n$ represent number of local views specifically set to 8.
The teacher parameters are updated via exponential moving average of the student weights  using: $\theta_{t} \leftarrow \lambda \theta_{t} +  (1 - \lambda \theta_{s})$ where $\theta_{t}$ and $\theta_{s}$ denote the parameters of teacher and student network respectively and, $\lambda$ follows the cosine schedule from 0.996 to 1 during training.  Further, we apply centering and sharpening operations to the teacher output. This way our method avoid any mode collapse similar to BYOL \cite{grill2020bootstrap} and converge to unique solution \cite{caron2021dino}.

Our self-supervised view prediction objective (Eq. \ref{eq: overall_loss}) on low resolution inputs induces locality in the Vision Transformer and encourages better intermediate feature representations which further aids during the fine-tuning stage on the same dataset.

\subsection{Self-supervised to Supervised Label Prediction}
\label{subsec: Self-supervised to Supervised Label Prediction}

We initialize a given model with weights learned via our self-supervised approach on the target dataset and then fine-tune the model on the same corresponding dataset. This is in contrast to existing practices of initializing the models with different initialization schemes \cite{NEURIPS2019_9015_pytorch} or ImageNet pre-trained weights. We transfer weights from the teacher network (Fig. \ref{fig: Our Training framework}) and replace the self-supervised MLP projection head with a randomly initialized MLP classifier. The model is then trained via supervised objective  as follows:

\begin{equation}\label{eq: ce}
\mathcal{L}_{CE} = - \sum_{i=1}^k y_i \cdot log(\mathcal{F}(\bm{x})_i),
\end{equation}

where $k$ is the output dimension of the final classifier and $y$ represents the one-hot encoded ground-truth. We note that teacher provides high quality target features during pre-training and hence prove useful for the fine-tuning stage \cite{caron2021dino}. Further, the ablation on the effect self-supervised weights is provided in Table \ref{tab: teacher-student-compare}.

\section{Experimental Protocols}
In this section, we discuss the experimental settings including dataset and training details , qualitative (Sec. \ref{subsec: Results}), and ablative analysis (Sec. \ref{subsec: Ablative Analysis}).

\noindent\textbf{Datasets:}
We validate our approach on five small-scale,  low-resolution datasets including  Tiny-Imagenet \cite{Le2015TinyImagenet}, CINIC10 \cite{darlow2018cinic}, CIFAR10, CIFAR100 \cite{Krizhevsky09cifar}, SVHN \cite{GOODFELLOW2013SVHN} and two finegrained datasets including Aircraft \cite{maji13fine-grained} and Cars \cite{KrauseStarkDengFei-Fei_3DRR2013}.  Details about the dataset size, sample resolution and the number of classes are provided in Table \ref{dataset-table}.

\noindent\textbf{Self-supervised Training Setup:}
We train all models with the Adam optimizer \cite{kingma2014adam} and a batch size of 256 via distributed learning over 4 Nvidia V100 32GB GPUs . We linearly ramp up the learning rate during the first 10 epochs using: $\textit{lr} = 0.0005 * \frac{\text{Batch size}}{256}$. After first 10 epochs, learning rate follows the cosine schedule \cite{ilya2016cosine}. We re-scale the student and teacher outputs using a temperature parameter \cite{devlin2018bert} which is set to 0.1  for the student network, while it follows a linear warm-up from 0.04 to 0.07 for the teacher network.

\noindent \textbf{Supervised Training Setup:} We use the training framework of \cite{lee2021vitsmall} for supervised learning and apply standard data augmentations for consistency.  Specifically, we use cutmix \cite{yun2019cutmix}, mixup \cite{yun2019cutmix}, auto-augment \cite{cubuk2018autoaugment}, and repeated augment \cite{cubuk2019randaug}. Further, we also use label smoothing \cite{szegedy2015labelsmooth}, stochastic depth \cite{huang2016stochastic}, and random erasing \cite{zhong2017erasing}.  We train all models for 100 epochs with a batch size of 256 on a single Nvidia V100 32GB GPU. We use Adam optimizer \cite{kingma2014adam} with a learning rate of 0.002 and learning decayed rate of 5e-2 with cosine scheduling. 

\vspace{-0.5em}
\begin{table}[!t]
    \begin{minipage}{.65\textwidth}
        \centering\small
        \scalebox{0.8}[0.8]{
        \begin{tabular}{|l|c|c|c|c|}
            \hline
            \rowcolor{Gray} 
            \textbf{Dataset}     &  \textbf{Train Size}   &  \textbf{Test Size}   &  \textbf{Dimensions}  &  \textbf{\# Classes}  \\
            \hline\hline
                Tiny-Imagenet \cite{Le2015TinyImagenet} & 100,000 &10,000 & 64x64    & 200 \\
                CIFAR10  \cite{Krizhevsky09cifar}   & 50,000 & 10,000  & 32x32    & 10      \\
                CIFAR100 \cite{Krizhevsky09cifar}    & 50,000 & 10,000 & 32x32    & 100  \\
                CINIC10 \cite{darlow2018cinic}   & 90,000  & 90,000 &32x32    & 10  \\
                SVHN \cite{GOODFELLOW2013SVHN}       & 73,257  & 26,032  & 32x32    & 10  \\
                Aircraft \cite{maji13fine-grained}       & 6,667  & 3533  & 224x224    & 102  \\
                Cars \cite{KrauseStarkDengFei-Fei_3DRR2013}       & 8,144  & 8,041  & 224x224    & 196  \\
            \hline
            \end{tabular}}
    \end{minipage}
    \hfill
    \begin{minipage}{0.34\textwidth}
    \caption{\small Details of datasets in terms of sample size and  resolution used in our experiments. We propose to learn Self-supervised initialization directly from small datasets with default resolution. 
    % This allows to train ViTs  on these datasets without any large-scale pre-training.
    }
    \label{dataset-table}
    \end{minipage}
    \vspace{-1.2em}
\end{table}

\begin{table}[!t]
  \begin{minipage}{1.0\textwidth}
   \centering \small
  \setlength{\tabcolsep}{5pt}
  \scalebox{0.85}[0.85]{
  \begin{tabular}{c|c|ccccc}
  \toprule
  \rowcolor{Gray} 
    \textbf{Model}   &  \textbf{Params(M)} &  \textbf{Tiny-Imagenet} &  \textbf{CIFAR10} &  \textbf{CIFAR100} &  \textbf{CINIC10} &  \textbf{SVHN} \\
    \midrule
    ResNet56 \cite{kaiming2015resnet}   & 0.9   & 56.51  & 94.65  &74.44   & 85.34  &97.61  \\ [2pt]
    ResNet110 \cite{kaiming2015resnet}  & 1.7   & 59.77 & 95.27  &76.18   & 86.81  &97.82   \\ [2pt]
    EfficientNet B0 \cite{Mingxing2019EfficientNet}    & 4.0    & 55.48  & 88.38  &61.64   & 75.64  &96.06     \\ [2pt]
    ResNet18  \cite{kaiming2015resnet}    & 11.6    & 53.32 & 90.44  &64.49   & 77.79  &96.78     \\ [2pt]
    \midrule

    ViT (scratch)     & 2.8   & 57.07  & 93.58  &73.81   & 83.73  &97.82    \\ [2pt]
     SL-ViT \cite{lee2021vitsmall}\textsubscript{(Arxiv'21)}    & 2.9   & 61.07  & 94.53  &76.92   & 84.48 &97.79    \\ [2pt]
     ViT-Drloc \cite{liu2021nipsvitsmall} \textsubscript{(NeurIPS'21)}    &3.15    &42.33   &81.00   &58.29   &71.50  &94.02    \\ [2pt]
    \rowcolor{LightCyan}
    ViT (Ours)   & 2.8   & \textbf{63.36}  & \textbf{96.41}  & \textbf{79.15}   & \textbf{86.91}  &\textbf{98.03}   \\ [2pt]
    % ViT-Pre (Ours) + SL \cite{lee2021vitsmall}    & 2.9   & 63.12  &96.19   & 79.01  & 86.95  & 98.01   
    % \\ [2pt]

    \midrule
    
    Swin (scratch)     & 7.1   & 60.05  & 93.97  &77.32   & 83.75  &97.83   \\ [2pt]
    SL-Swin \cite{lee2021vitsmall} \textsubscript{(Arxiv'21)}   & 10.2   & 64.95  & 94.93  &79.99   & 87.22 &97.92       \\ [2pt]
    Swin-Drloc \cite{liu2021nipsvitsmall} \textsubscript{(NeurIPS'21)}    & 7.7   &48.66   &86.07  &65.32    &77.25   & 95.77      \\ [2pt]
    \rowcolor{LightCyan}
    Swin (Ours)     & 7.1    & \textbf{65.13}  & \textbf{96.18} & \textbf{80.95}   & \textbf{87.84}  & \textbf{98.01}       \\ [2pt]
    % Swin (Ours) + SL \cite{lee2021vitsmall}    & 10.2   & 66.72  & 96.84  & 80.44  & 88.74 & 97.97 
    % \\ [2pt]
    
    \midrule
    
    CaiT (scratch)     & 7.7    & 64.37  & 94.91  &76.89   & 85.44  &98.13   \\ [2pt]
    SL-CaiT \cite{lee2021vitsmall} \textsubscript{(Arxiv'21)}    & 9.2   & 67.18  & 95.81  &80.32  & 86.97  & \textbf{98.28}       \\ [2pt]
    CaiT-DRLoc \cite{liu2021nipsvitsmall} \textsubscript{(NeurIPS'21)}    &8.5  &45.95    &82.20   &56.32  &73.85    & 19.59       \\ [2pt]
    \rowcolor{LightCyan}
    CaiT (Ours)    & 7.7    & \textbf{67.46}  & \textbf{96.42} & \textbf{80.79}   & \textbf{88.27}  & 98.18    \\ [2pt]
    % CaiT (Ours) + SL \cite{lee2021vitsmall}    & 10.2   & 68.40  & 96.51  & 81.63  & 88.54 & 98.14 
    % \\ [2pt]
    
  \bottomrule
  \end{tabular}}
  \begin{minipage}{1.0\textwidth}
  \centering
  \vspace{0.3em}
       \caption{\small Our approach performs favorably well against different ViT baselines \cite{lee2021vitsmall, liu2021nipsvitsmall} as well as CNNs without adding any additional parameters or requiring changes to architecture or loss functions. Note that all methods are trained on the original input resolution as provided in Table \ref{tab: architectures}.}
       \label{table: main results}
  \end{minipage}
  \end{minipage}
 \vspace{-1.2em}
\end{table}

\begin{table}[!t]
  \begin{minipage}{1.0\textwidth}
   \centering \small
  \setlength{\tabcolsep}{3pt}
  \scalebox{0.82}[0.82]{
  \begin{tabular}{c|c|c|c|c|c|ccc}
  \toprule
  \rowcolor{Gray} 
    \textbf{Model} &  \textbf{Method} &  \textbf{Input Resolution} &  \textbf{Patch-size} & \textbf{No. of Tokens} & \textbf{Params(M)}  &  \textbf{CIFAR10} &  \textbf{CIFAR100} &  \textbf{SVHN}\\
    \midrule
    Swin  & Drloc \cite{liu2021nipsvitsmall} & 224x224 &4 & 3136 &  28  & 83.89 &66.23 &94.23\\
    
    Swin (Ours)  & Drloc \cite{liu2021nipsvitsmall} & 32x32 &2 & 64 & 7.7 & 86.07 & 65.32 & 95.77\\
    \rowcolor{LightCyan}
     Swin (Ours)  & Ours  & 224x224 &4 & 3136  & 7.7 & 92.04 & 73.46 & 96.86\\
  \bottomrule
  \end{tabular}}
  \end{minipage}
  \begin{minipage}{1.0\textwidth}
  \centering 
  \vspace{0.3em}
  \caption{\small In comparison to \cite{liu2021nipsvitsmall}, the performance of our approach improves significantly on higher resolution.  Thus our approach proves effective on both low as well as high input resolutions. }
%   \vspace{-1em}
%   \caption{\small We observe that \cite{liu2021nipsvitsmall} is sensitive to the input resolution and data distribution. In contrast, our approach proves effective on low as well as high input resolutions }\vspace{-1em}
 \label{tab: drloc-compare}
 \end{minipage}
 \vspace{-1.8em}
\end{table}

 \begin{table}[!t]
  \begin{minipage}{0.4\linewidth}
   \centering \small
  \setlength{\tabcolsep}{18pt}
  \scalebox{0.7}[0.7]{
  \begin{tabular}{c|cc}
  \toprule
  \rowcolor{Gray} 
    \textbf{Method}  & \textbf{Aircraft} &\textbf{Cars} \\
    \midrule
   ViT-Drloc [38]   & 10.40 & 13.82 \\
   \rowcolor{LightCyan}
   ViT (Ours)  &\textbf{66.04} & \textbf{43.89} \\
  \bottomrule
  \end{tabular}}
  \end{minipage}
   \hfill
  \begin{minipage}{0.55\linewidth}
  \caption{\small Our approach outperforms the existing SOTA approach on the finegrained datasets (Top-1 accuracy).}  
 \label{tab: finegrained}
  \end{minipage}
     \vspace{-0.5em}
\end{table}

\subsection{Results}
\label{subsec: Results}

\textbf{Generalization:} We observe generalization of different methods with a comparative analaysis presented in Table \ref{table: main results}  across 3 different ViT architectures (Table \ref{tab: architectures}). We keep a patch size of 8  for Tiny-Imagenet to generate 16 number of input tokens for ViT and CaiT architectures. We reduce the patch size to 4 so that the resultant  number of tokens become 64 for all other datasets such as CIFAR, SVHN, and CINIC10. Similarly, for Swin architecture, we use a patch size of 4 for Tiny-Imagenet to obtain 64 tokens, while for other datasets, we use a patch size of 2 which produces 256 number of input tokens. We consistently follow these architectural settings for all the baselines (Table \ref{table: main results}). Our approach consistently performs better as compared to recent state-of-the-art methods (\cite{lee2021vitsmall, liu2021nipsvitsmall}) for ViTs training on small-scale datasets (Table \ref{table: main results}). Particularly, we observe a significant gain for the difficult cases where the ratio of number of classes vs. input samples is higher \eg CIFAR100 and Tiny-ImageNet (Table \ref{table: main results}). In this manner, our approach paves the way to adopt ViTs to small datasets that also outperforms CNN based models. The effect of our self-supervised wight initialization on convolutional networks is provided in Appendix \ref{sec:convenet-self-supervised}.

\noindent\textbf{Performance on fine-grained datasets:}
We test our approach on two fine-grained datasets such as Aircarft and Cars, and present the top-1 accuracy results in Table \ref{tab: finegrained}. We observe that our proposed approach performs well on fine-grained datasets compared to  \cite{liu2021nipsvitsmall}. 
% which struggles to train on these datasets.

\noindent\textbf{Robustness to Input Resolution and Patch Sizes:} 
%For small datasets, we observe that ViT is sensitive to the input resolution and patch size to create input tokens. 
A recent method \cite{liu2021nipsvitsmall} projects the input samples to higher resolution to train Vision Transformer \eg input resolution of 32x32 for CIFAR is re-scaled to 224x224 during training. This significantly increases the number of input tokens and hence the quadratic complexity within self-attention (Table. \ref{tab: drloc-compare}). In comparison, our approach successfully trains ViTs on low resolution inputs while being computationally efficient. Our method scales well on high resolution inputs and outperforms \cite{liu2021nipsvitsmall} by notable margins (Table \ref{tab: drloc-compare}).

\noindent\textbf{Robustness to Natural Corruptions:}  We analyse the mean corruption error on CIFAR10 and CIFAR100 in Table \ref{tab: corruptions-table}. Our training approach increases the model robustness against 18 natural corruptions such as fog, rain, noise, and blur, etc. \cite{hendrycks2018benchmarking}.

\noindent\textbf{Attention to Salient Regions:} We visualize self-attention in Fig. \ref{fig:atten-maps}.  The attention scores of the class token computed across the attention heads for the last ViT block is projected  onto the unseen test samples of Tiny-ImageNet. Our proposed approach is able to capture the shape of the salient objects more efficiently with minimal or no attention to the background as compared to the baseline approaches where the attention is more spread out in the background and they completely fail to capture the shape of the salient object in the image.

\begin{table}[!t]
 \centering \small
  \setlength{\tabcolsep}{5pt}
  \scalebox{0.9}[0.9]{
  \begin{tabular}{c|ccacca}
  
  \toprule
  \rowcolor{Gray} 
    \textbf{Data}   & \textbf{ ViT (scratch)}  & \textbf{SL-ViT} & \textbf{ViT (Ours)} & \textbf{Swin (scratch)} & \textbf{SL-Swin }& \textbf{ Swin (Ours)}\\
    \midrule
  CIFAR10 & 39.93  & 26.42 & \textbf{26.01} &  36.13 & 26.28 & \textbf{25.38}  \\
  CIFAR100 & 65.04 & 48.56 & \textbf{48.10} & 53.83 & 47.29 & \textbf{45.10}\\
  \bottomrule
  \end{tabular}}
  \vspace{1em}
    \caption{\small Mean corruption error (\emph{lower is better}) is reported against 18 natural corruptions \cite{hendrycks2018benchmarking}. Our training method improves model robustness against such real world corruptions such as fog, rain, etc.}

 \label{tab: corruptions-table}
 \vspace{-1em}
\end{table}

\begin{table}
\begin{minipage}{0.40\textwidth}
  \centering
  \includegraphics[width=\linewidth, scale=0.5]{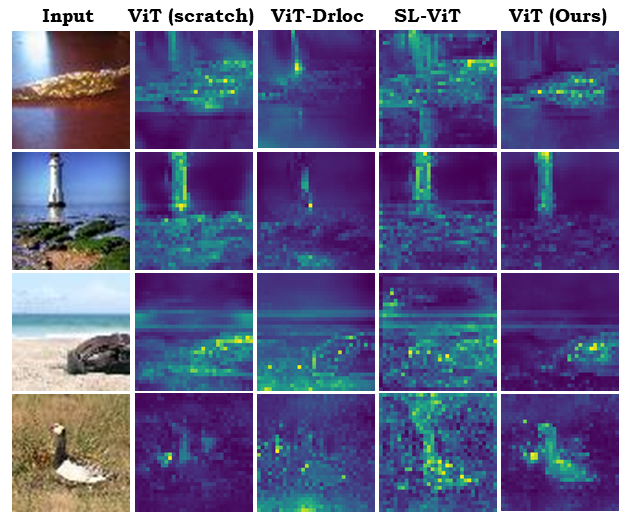}
  \vspace{0.1em}
  \captionof{figure}{\small Our approach captures  the salient objects in the image in comparison to baseline methods for which the attention is dispersed in the background.}
  \label{fig:atten-maps}
\end{minipage}
\hfill
\begin{minipage}{0.58\textwidth}
 % lelf lower right up 
  \includegraphics[width=\linewidth, trim= 0mm 0mm 2.5mm 0mm, clip]{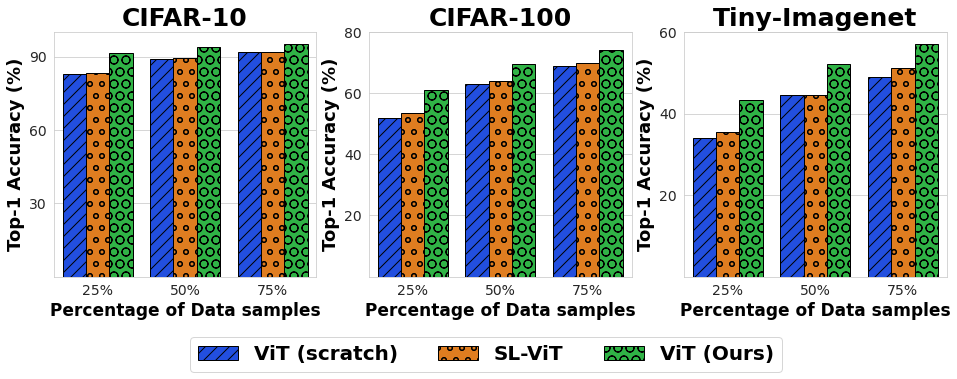}
    \vspace{0.5em}
  \captionof{figure}{\small We demonstrate the efficiency of our approach by varying the training samples of CIFAR10, CIFAR100 and Tiny-Imagenet. Specifically we train the models with 25\%, 50\% and 75\% of training data. \emph{Our approach consistently performs better even with limited training data.}}
  \label{fig:train_efficiency}

 \end{minipage}
 \vspace{-1em}
\end{table}

\subsection{Ablative Analysis}
\label{subsec: Ablative Analysis}

\noindent \textbf{Effect of Data Size on Self-supervised Weight Initialization:} We study the effect of data size on our self-supervised learning for weight initialization (Fig. \ref{fig:train_efficiency}). We train the ViT models \cite{touvron2020deit} on 25\%, 50\% and 75\% of the training samples across 3 datasets: CIFAR10, CIFAR100 and Tiny-Imagenet datasets. In case of CIFAR10, our approach achieves more than 90\% top-1 accuracy with just 25\% of data and outperforms other approaches with notable margins. We observe a similar trend with CIFAR100 and Tiny-Imagenet datasets.%, where our proposed approach surpasses the baseline methods.

\noindent \textbf{Effect of Local-Global Crop Ratio:} We generate local and global views by randomly cropping certain regions from the original input image. The cropped area of each generated view is chosen from a specified range of values w.r.t the original input size. We  analyse the impact of the range of aspect ratios for local and global views w.r.t the original input size in Table \ref{local-global-table-cif}. The original input size of Tiny-Imagenet is 2x times greater than the other datasets used in our experiments, therefore, we use a modified range of local-global aspect ratios as shown in Table \ref{local-global-table-cif} (right). We observe that range of aspect ratios between (0.2, 0.4) for local view and  (0.5, 0.1) for global view  works well for the Tiny-ImageNet. Similarly, for the other relatively lower-resolution datasets, the optimal aspect ratios are in the range of (0.2, 0.5) and (0.7, 1.0)  for the local and global views, respectively (Table \ref{local-global-table-cif}).% is chosen  in the range of (0.2, 0.5) and  global view aspect ratio is set to a range of (0.7, 1.0).

\noindent \textbf{Effect of Self-supervised MLP Dimensions:} Table \ref{head-dim-table} shows the effect of the output head dimension of our self-supervised projection MLP on the model generalization during supervised fine-tuning stage. We fix the local-global aspect ratio to their optimal values and ablate over a range of MLP head dimensions. Based on the top-1 accuracy results on train set (Table \ref{head-dim-table}), we choose the dimension of size 1024 for all our experiments.

\noindent \textbf{Effect of Teacher Vs. Student Weights Transfer:} We compare the performance of the ViT initialized with our self-supervised weights from student and teacher networks (Table \ref{tab: teacher-student-compare}). We observe higher generalization (top-1 accuracy) with the teacher weights that corroborates our strategy of choosing teacher rather than student weights for the supervised training stage.

\begin{table}[!t]
  \begin{minipage}{0.3\textwidth}
   \centering \small
  \setlength{\tabcolsep}{3pt}
  \scalebox{0.7}[0.7]{
  \begin{tabular}{c|c|cc}
  \toprule
  \rowcolor{Gray} 
    \textbf{Model}  & \textbf{Weights} &  \textbf{CIFAR100} & \textbf{Tiny-Imagenet} \\
    \midrule
    ViT  & Student & 77.27 &  61.02\\
    ViT  & Teacher & \textbf{79.15} & \textbf{63.36}\\
  \bottomrule
  \end{tabular}}
  \end{minipage}
  \hfill
  \begin{minipage}{0.63\textwidth}
  \caption{\small Self-supervised teacher weights transfer well as compared to the student. This finding is consistent with \cite{caron2021dino}.}
 \label{tab: teacher-student-compare}
 \end{minipage}
  \vspace{-1em}
\end{table}

\begin{table}[!t]
%\vspace{-0.3cm}
\begin{minipage}{0.5\textwidth}
   \centering \small
  \setlength{\tabcolsep}{4pt}
  \scalebox{0.8}[0.8]{
\begin{tabular}{llcccc}
\toprule
  \rowcolor{Gray} 
    \multirow{-1}{*}{\textbf{Local View}}&  \multicolumn{1}{c}{\multirow{-1}{*}{\textbf{Global View}}} &
    \multicolumn{2}{c}{\textbf{CIFAR10}} &
    \multicolumn{2}{c}{\textbf{CIFAR100}}\\
    \cline{3-6}
    \rowcolor{Gray} 
    & & \multicolumn{1}{c}{\textbf{ViT}} & \multicolumn{1}{c|}{\textbf{Swin}} & \multicolumn{1}{c}{\textbf{ViT}} & \multicolumn{1}{c}{\textbf{Swin}}\\
\midrule
    (0.1 , 0.4)    & \multicolumn{1}{c|}{(0.4 , 1.0)}  &      77.36 & \multicolumn{1}{c|}{67.82} & 79.52 &74.10\\
    (0.15 , 0.4)& \multicolumn{1}{c|}{(0.4 , 1.0)}   &      78.90 & \multicolumn{1}{c|}{68.12} &78.14 &74.06 \\  
    (0.2 , 0.5) & \multicolumn{1}{c|}{(0.5 , 1.0)}     &     77.48 & \multicolumn{1}{c|}{67.90} &79.53 &74.05 \\  
    (0.2 , 0.5) & \multicolumn{1}{c|}{(0.6 , 1.0)}   &      78.87 & \multicolumn{1}{c|}{65.52} &79.51 &74.09 \\
    \textbf{(0.2 , 0.5)} & \multicolumn{1}{c|}{(\textbf{0.7 , 1.0)}}   &     \textbf{ 79.19 }& \multicolumn{1}{c|}{\textbf{71.64}} &\textbf{79.78} &\textbf{74.67} \\
\bottomrule
\end{tabular}}
\end{minipage}
\hfill
\begin{minipage}{0.45\textwidth}
\centering \small 
  \setlength{\tabcolsep}{9pt}
  \scalebox{0.8}[0.8]{
    \begin{tabular}{llcc}
      \toprule
      \rowcolor{Gray} 
        \multirow{-1}{*}{\textbf{Local View}}&  \multicolumn{1}{c}{\multirow{-1}{*}{\textbf{Global View}}} &
        \multicolumn{2}{c}{\textbf{Tiny-ImageNet}}\\
        \cline{3-4}
        \rowcolor{Gray} 
        & & \multicolumn{1}{c}{\textbf{ViT}} & \multicolumn{1}{c}{\textbf{Swin}}\\
        \midrule
        (0.1 , 0.3) & (0.3 , 1.0)  & 64.37 & 74.33\\
        (0.15 , 0.45) & (0.45 , 1.0)  & 62.02 & 74.09\\
        % (0.2 , 0.4) & (0.4 , 1.0)  & 62.03 & 74.06\\
        % \rowcolor{LightCyan}
        \textbf{(0.2 , 0.4)} & \textbf{(0.5 , 1.0)}  & \textbf{64.82} & \textbf{75.13} \\
        (0.2 , 0.4) & (0.6 , 1.0)  & 62.03 & 57.33\\
        % (0.2 , 0.5) & (0.5 , 1.0)  & 62.08 & 74.02\\
        (0.3 , 0.5) & (0.5 , 1.0)  & 61.69 & 73.92\\
        
      \bottomrule
      \end{tabular}
}
\end{minipage}
\vspace{0.5em}
\caption{\small We demonstrate the impact of local-global crop ratios chosen during self-supervised training on the Top-1 train accuracy scores of CIFAR10 and CIFAR100 (left), and Tiny-Imagenet (right).}
\label{local-global-table-cif}  
\vspace{-1.5em}
\end{table}

\begin{table}[!t]
\centering
\begin{minipage}{0.55\textwidth}

  \centering
    \setlength{\tabcolsep}{6.5pt}
  \scalebox{0.7}[0.7]{
\begin{tabular}{cccccccc}
\toprule
  \rowcolor{Gray} 
    \multirow{-1}{*}{\textbf{SSL Head}}&
    \multicolumn{2}{c}{\textbf{CIFAR10}} &
    \multicolumn{2}{c}{\textbf{CIFAR100}} &
    \multicolumn{2}{c}{\textbf{Tiny-Imagenet}}\\
    \cline{2-7}
     \rowcolor{Gray} 
    & \textbf{ViT} & \multicolumn{1}{c|}{\textbf{Swin}}
        & \multicolumn{1}{c}{\textbf{ViT}} & \multicolumn{1}{c|}{\textbf{Swin}}
        & \multicolumn{1}{c}{\textbf{ViT}} & \multicolumn{1}{c}{\textbf{Swin}}\\
\midrule
    \multicolumn{1}{c|}{512}    & 78.77  & \multicolumn{1}{c|}{79.46} & 71.63 & \multicolumn{1}{c|}{74.04} & 66.84 & 74.10\\
    \multicolumn{1}{c|}{\textbf{1024}}    & \textbf{79.19}  & \multicolumn{1}{c|}{\textbf{79.78}} & \textbf{71.64} & \multicolumn{1}{c|}{\textbf{74.67}} &\textbf{ 65.82} & \textbf{75.13}\\
    \multicolumn{1}{c|}{2048}    & 78.83  & \multicolumn{1}{c|}{79.48} & 71.15 & \multicolumn{1}{c|}{73.87} & 65.05 & 73.95\\
    \multicolumn{1}{c|}{4096}    & 78.92  & \multicolumn{1}{c|}{79.50} & 71.03 & \multicolumn{1}{c|}{74.21} & 66.48 & 74.03\\

\bottomrule
\end{tabular}
}
\end{minipage}
\hfill
\begin{minipage}{0.44\textwidth}
 \caption{\small \emph{Effect of Self-supervised MLP Projection Head:} We observe that MLP head dimension of 1024 gives better overall results on train set across 3 datasets using ViT and Swin architectures.}
\label{head-dim-table}
\end{minipage}
\vspace{-0.7em}
\end{table}

\noindent \textbf{Performance comparison with
self-supervised learning based CNN:} 
We provide a comparison of ViT with ResNet18 ( 2.8 vs.
11.6 million parameters) in Table \ref{tab: resnet18-vs-ours-ssl}. ViT’s performance improves significantly in comparison to self-supervised CNN.
This finding is consistent with \cite{caron2021dino}. In addition, we compare
our proposed self-supervised approach with different contrastive self-supervised methods \cite{ting2020simclr, chen2021mocov3} that are mainly studied for CNNs (Table \ref{tab: ssl-for-cnn}). Our method provides SOTA results
for ViTs in comparison to self-supervised CNN.

\begin{table}[!t]
  \begin{minipage}{0.55\linewidth}
   \centering \small
  \setlength{\tabcolsep}{11pt}
  \scalebox{0.7}[0.7]{
  \begin{tabular}{cc|cc}
  \toprule
  \rowcolor{Gray} 
    \textbf{Model}   & \textbf{Initialization} & \textbf{CIFAR100} & \textbf{Tiny-Imagenet}  \\
    \midrule
   ResNet-18  & SS (ours)  &65.00  &53.48\\
   ViT  & SS (ours)  &\textbf{79.15}  &\textbf{63.36}\\
  \bottomrule
  \end{tabular}}
  \end{minipage}
   \hfill
  \begin{minipage}{0.45\linewidth}
  %   \vspace{1em}
  \caption{\small Top-1 accuracy comparison of self-supervised (SS) trained CNN with Ours. }
 \label{tab: resnet18-vs-ours-ssl}
 
  \end{minipage}
\vspace{-0.8em}
\end{table}

 \begin{table}[!t]
  \begin{minipage}{0.55\linewidth}
   \centering \small
  \setlength{\tabcolsep}{11pt}
  \scalebox{0.7}[0.7]{
  \begin{tabular}{c|ccc}
  \toprule
  \rowcolor{Gray} 
    \textbf{Method} & \textbf{Tiny-Imagenet} &\textbf{CIFAR10} &\textbf{CIFAR100} \\
    \midrule
   SimCLR \cite{ting2020simclr}    &58.87  & 93.50 & 74.77 \\
   MOCO-V3 \cite{chen2021mocov3} &52.39  &93.95  & 72.22 \\
    % \rowcolor{LightCyan}
%   SimMIM   &50.71  &93.90  & 71.26\\
   Ours  & \textbf{63.36} &\textbf{96.41}  & \textbf{79.15}\\
  \bottomrule
  \end{tabular}}
  \end{minipage}
   \hfill
  \begin{minipage}{0.45\linewidth}
  \caption{\small Comparison of other existing self-supervised learning techniques with ours using basic ViT architecture.}
 \label{tab: ssl-for-cnn}
  \end{minipage}
   \vspace{-0.8em}
\end{table}

 \begin{table}[!t]
  \begin{minipage}{0.55\linewidth}
   \centering \small
  \setlength{\tabcolsep}{13pt}
  \scalebox{0.78}[0.78]{
  \begin{tabular}{cc|cc}
  \toprule
  \rowcolor{Gray} 
    \textbf{Method}  & \textbf{Epochs} & \textbf{CIFAR100} \\
    \midrule
   ViT-Drloc \cite{liu2021nipsvitsmall}    & 600 (supervised) & 68.29 \\
   ViT (Ours)  & 200(self-supervised) & \textbf{76.08} \\
     & + 100(supervised) & \\
    % \rowcolor{LightCyan}
  \bottomrule
  \end{tabular}}
  \end{minipage}
   \hfill
  \begin{minipage}{0.45\linewidth}
  \caption{\small Our approach is efficient in terms of epochs used and outperforms the current approach in terms of Top-1 accuracy.}  
 \label{tab: epoch-compare}
  \end{minipage}
   \vspace{-0.8em}
\end{table}

\begin{table}[!t]
  \begin{minipage}{0.55\linewidth}
   \centering \small
  \setlength{\tabcolsep}{14pt}
  \scalebox{0.67}[0.67]{
  \begin{tabular}{c|ccc}
  \toprule
  \rowcolor{Gray} 
    \textbf{Self-supervised MLP layers}  &\textbf{CIFAR100} &\textbf{Tiny-Imagenet} \\
    \midrule
   1-Layer   &76.69  & 60.54  \\
   3-Layer (Ours)  &\textbf{79.15}  & \textbf{63.36} \\
    % \rowcolor{LightCyan}
  \bottomrule
  \end{tabular}}
  \end{minipage}
   \hfill
  \begin{minipage}{0.45\linewidth}
  \caption{\small Top-1 accuracy comparison of 3-Layer MLP with 1-Layer MLP training}
 \label{tab: 1layer-vs-3layer}
  \end{minipage}
 \vspace{-0.8em}
\end{table}

\begin{table}[!t]
  \begin{minipage}{0.55\linewidth}
   \centering \small
  \setlength{\tabcolsep}{14pt}
  \scalebox{0.85}[0.85]{
  \begin{tabular}{c|ccc}
  \toprule
  \rowcolor{Gray} 
    \textbf{Head size}  &\textbf{CIFAR100} &\textbf{Tiny-Imagenet} \\
    \midrule
   65536 \cite{caron2021dino}    &77.42  & 60.77  \\
   1024 (Ours)  &\textbf{79.15}  & \textbf{63.36} \\
    % \rowcolor{LightCyan}
  \bottomrule
  \end{tabular}}
  \end{minipage}
   \hfill
  \begin{minipage}{0.45\linewidth}
  \caption{\small Top-1 accuracy comparison of the size of projection head used during self-supervised training}
 \label{tab: head-compare}
  \end{minipage}
   \vspace{-0.8em}
\end{table}

\noindent \textbf{Efficiency in terms of epochs:}  We further observe that our method trained for 300 epochs (200 for self-supervised view matching and 100 for
supervised label prediction) outperforms the current SOTA
approach \cite{liu2021nipsvitsmall} trained for even 600 epochs (Table \ref{tab: epoch-compare}).

\noindent \textbf{Self-supervised MLP layers:} We modify the self-supervised projection
MLP which reduces complexity and increases generalization as shown in Table \ref{tab: 1layer-vs-3layer}.

\noindent \textbf{Analysis of MLP Head:}  We observe that larger the size
of MLP such as 65536 \cite{caron2021dino}, the lower is the performance on
small-scale datasets (Table \ref{head-dim-table} and Table \ref{tab: head-compare}). This is because
the large size of MLP head might result in overfitting
the features of low resolution views.

\vspace{-1em}
\section{Conclusion}
In this work, we introduce an effective strategy to train Vision Transformers on small-scale low-resolution datasets without large-scale pre-training. %While CNN's have inherent inductive biases, ViTs on the other hand lack such properties due to which these architectures struggle on small datasets. 
 %Specifically we learn the self-supervised weights by employing view matching objective.  
We propose to learn self-supervised inductive biases directly from the small-scale datasets. We initialize the network with the weights learned through self-supervision and fine-tune it on the same dataset during the supervised training.
We show through extensive experiments that our method can serve as a better initialization scheme and hence allows to train ViTs from scratch on small datasets while performs favorably well w.r.t the existing state-of-the-art methods. Further, our approach can be used in a plug-and-play manner for different ViT designs and training frameworks without any modifications to the architectures or loss functions.

% We believe that the simplicity and effectiveness of our approach makes it easy to reproduce and train. In addition to this, being architecture agnostic, we can plug and play any ViT architecture in the training framework, thus making the framework flexible to any modifications. We hope that our work inspires further research in this domain.

\bibliography{references}

\newpage
\appendix

\noindent\begin{huge} \textbf{Appendix} \vspace{4mm} \end{huge}

We extensively study the effect of different ViT architectural attributes including network depth by adding more self-attention blocks sequentially and attention heads within each self-attention block in Appendix \ref{sec:vit-vs-vit-full}. The effect of our proposed self-supervised approach on CNNs is provided in Appendix \ref{sec:convenet-self-supervised}. Finally, in Appendix \ref{sec:appendix-ssl-attn-maps}, we provide visual examples capturing salient region in the input samples.
% \section{Effect of Changing Depth and Attention heads on ViT's Performance}
\section{Exploring ViT Attributes: Depth, Attention-Heads}
\label{sec:vit-vs-vit-full}
We modify the attributes of ViT architecture and observe the effect on model generalization (top-1 accuracy \%) across CIFAR100 and Tiny-Imagenet datasets (Table \ref{tab: vit-full-vs-ours}) . Specifically, we vary the depth and attention heads to study the relation between ViT parameter complexity and its generalization. Our analysis highlights the following insights: 
% This modification is motivated from the work of \cite{raghu2021vitcnn} who empirically show that lower layers of ViT contain some level of locality which diminishes with the higher layers. Furthermore, the attention distance increase with the increase in the ViT depth. Therefore, to retain some locality in the network, we reduce the depth of ViT and increase the attention heads. 
%From the results in Table \ref{tab: vit-full-vs-ours}, we conclude that our choice of attributes for the ViT architecture are optimal as it provides best top-1 accuracy on CIFAR100 and Tiny-Imagenet datasets.
\tcblack{\small{1}} For a given training method, the performance of model improves as we increase the number of self-attention blocks (\eg six to nine), however, we observe the decrease in generalization by further increasing the self-attention blocks (\eg at 12). This finding is consistent with \cite{raghu2021vitcnn} that shows the reduced locality (inductive bias) within ViTs with a higher number of self-attention layers which adds to further difficulty to ViT optimization. \tcblack{\small{2}} Increasing number of heads within self-attention bring more diversity during training and lead to better results. \tcblack{\small{3}} Our approach outperforms the baseline methods in all the given settings, validating the necessity of self-supervised weight initialization during supervised learning.
% Our choice of attributes for ViT architecture are directly influenced from the work of \cite{raghu2021vitcnn} who empirically show that lower layers of ViT contain some level of locality which diminishes with the higher layers. Furthermore, the attention distance increase with the increase in the ViT depth. Therefore, to retain some locality in the network, we reduce the depth of ViT and increase the attention heads.

\begin{table}[!h]
\begin{minipage}{\textwidth}
   \centering \small
  \setlength{\tabcolsep}{4pt}
  \scalebox{0.6}[0.6]{
\begin{tabular}{lllllcccccccc}
\toprule
  \rowcolor{Gray} 
    \multirow{-1}{*}{Depth}&  \multicolumn{1}{c}{\multirow{-1}{*}{Patch-size}} &
    \multicolumn{1}{c}{Token-D}&
    \multicolumn{1}{c}{Heads} &
    \multicolumn{1}{c}{MLP-ratio}&
    \multicolumn{2}{c}{ViT-Scratch}&
    \multicolumn{2}{c}{ViT-Drloc}&
    \multicolumn{2}{c}{SL-ViT}&
    \multicolumn{2}{c}{ViT (Ours)}\\
    \cline{6-13}
    \rowcolor{Gray} 
    & & & & & \multicolumn{1}{c}{CIFAR100} & \multicolumn{1}{c|}{T-Imagenet} & \multicolumn{1}{c}{CIFAR100} & \multicolumn{1}{c|}{T-Imagenet}&
    \multicolumn{1}{c}{CIFAR100} & \multicolumn{1}{c|}{T-Imagenet}&
    \multicolumn{1}{c}{CIFAR100} & \multicolumn{1}{c}{T-Imagenet}
    
    \\
\midrule
      
      \multicolumn{1}{c}{6}  &[4,8] &192 &12 & \multicolumn{1}{c|}{2}      & 69.75 & \multicolumn{1}{c|}{53.44} & 51.97 & \multicolumn{1}{c|}{39.69} & 71.93 & \multicolumn{1}{c|}{54.94} & \textbf{75.14 }& \multicolumn{1}{c}{\textbf{59.15}} \\
      \multicolumn{1}{c}{6}  &[4,8] &192 &6 & \multicolumn{1}{c|}{2}      & 68.76 & \multicolumn{1}{c|}{52.25} & 53.12 & \multicolumn{1}{c|}{39.12} & 71.01 & \multicolumn{1}{c|}{53.68} & \textbf{72.92} & \multicolumn{1}{c}{\textbf{55.83}} \\
      \multicolumn{1}{c}{9}  &[4,8] &192 &12 & \multicolumn{1}{c|}{2}      & 73.81 & \multicolumn{1}{c|}{57.07} & 58.29 & \multicolumn{1}{c|}{42.33} & 76.92 & \multicolumn{1}{c|}{61.00} & \textbf{79.15} & \multicolumn{1}{c}{\textbf{63.36}} \\
      \multicolumn{1}{c}{9}  &[4,8] &192 &6 & \multicolumn{1}{c|}{2}      & 69.88 & \multicolumn{1}{c|}{53.56} & 55.50 & \multicolumn{1}{c|}{45.93} & 72.01 & \multicolumn{1}{c|}{54.63} & \textbf{73.59} & \multicolumn{1}{c}{\textbf{58.18}} \\
      \multicolumn{1}{c}{12}  &[4,8] &192 &3 & \multicolumn{1}{c|}{4}      & 68.09 & \multicolumn{1}{c|}{51.26} &  57.18& \multicolumn{1}{c|}{43.50} & 72.14 & \multicolumn{1}{c|}{52.98} & \textbf{68.88} & \multicolumn{1}{c}{\textbf{52.89}} \\
      \multicolumn{1}{c}{12}  &[4,8] &192 &12 & \multicolumn{1}{c|}{2}      & 71.23 & \multicolumn{1}{c|}{54.55} & 56.50 & \multicolumn{1}{c|}{45.71} & 74.04 & \multicolumn{1}{c|}{56.13} & \textbf{77.22} & \multicolumn{1}{c}{\textbf{56.41}} \\ 
      \multicolumn{1}{c}{12}  &[4,8] &192 &6 & \multicolumn{1}{c|}{2}      & 70.57 & \multicolumn{1}{c|}{53.42} & 58.58 & \multicolumn{1}{c|}{46.78} & 73.23 & \multicolumn{1}{c|}{55.38} & \textbf{73.93} & \multicolumn{1}{c}{\textbf{58.46}} \\ 
\bottomrule
\end{tabular}}
\end{minipage}
  \vspace{1em}
    \caption{\small The effect of ViT architectural attributes on its generalization. Our proposed approach performs favorably well in all the considered settings.} 
%   \caption{\small We report the top-1 accuracy of ViTs with different architectural attributes. We observe that our choice of depth and attention heads for the ViT are optimal as we get a higher top-1 accuracy with these attributes across CIFAR100 and Tiny-Imagenet datasets.} 
  %We observe that our ViT architecture with lower depth and more attention heads performs better than the original ViT-Tiny architecture with higher depth and lower attention heads. }
 \label{tab: vit-full-vs-ours}
 \vspace{-1em}
\end{table}

% \begin{table}[!h]
%   \begin{minipage}{1.0\textwidth}
%   \centering \small
%   \setlength{\tabcolsep}{3pt}
%   \scalebox{0.8}[0.8]{
%   \begin{tabular}{cccccc|cc}
%   \toprule
%   \rowcolor{Gray} 
%     \textbf{ViT Attributes $\rightarrow$}   & \textbf{Depth} & \textbf{Patch-size} & \textbf{Token Dimension} & \textbf{Heads} & \textbf{MLP-ratio} & \textbf{CIFAR100} & \textbf{Tiny-Imagenet} \\
%     \midrule
%     & 12 & [4,8] & 192 & 3 & 4 &68.88 &52.89\\
%     & 12 & [4,8] & 192 & 12 & 2 &77.22 &56.41 \\
%     & 12 & [4,8] & 192 & 6 & 2 &73.93 &58.46 \\
%     % \rowcolor{LightCyan}
%     & \textbf{9} & [4,8] & 192 & \textbf{12} & 2 &\textbf{79.15} &\textbf{63.36}\\
%     & 9 & [4,8] & 192 & 6 & 2 &73.59 &58.18\\
%     & 6 & [4,8] & 192 & 12 & 2 &75.14 &59.15 \\
%     & 6 & [4,8] & 192 & 6 & 2 &72.92 & 55.83\\

%   \bottomrule
%   \end{tabular}}
%   \end{minipage}
%   \vspace{2em}
%   \caption{\small We report the top-1 accuracy of ViTs with different architectural attributes. We observe that our choice of depth and attention heads for the ViT are optimal as we get a higher top-1 accuracy with these attributes across CIFAR100 and Tiny-Imagenet datasets.} 
%   %We observe that our ViT architecture with lower depth and more attention heads performs better than the original ViT-Tiny architecture with higher depth and lower attention heads. }
%  \label{tab: vit-full-vs-ours}
%  \vspace{-1em}
% \end{table}

\section{Self-supervised Weight Initialization for CNNs}
\label{sec:convenet-self-supervised}
\noindent Our self-supervised weight initialization strategy improves the performance of Vision Transformers. We study its effect on CNNs' performance in Table \ref{tab: resnet18-self-supervised-weight}. Specifically we pre-train a ResNet-18 model on Tiny-Imagenet and CIFAR100 datasets with our self-supervised view prediction objective and fine-tune it on the same datasets using supervised training framework (Sec. \ref{subsec: Self-supervised to Supervised Label Prediction} in main paper). We observe slight improvements in model performance as shown in Table \ref{tab: resnet18-self-supervised-weight}. 
%suggesting that self-supervised pre-training is less effective in CNN architectures . 
This shows that the presence of inherent inductive biases ease the CNN optimization with non-learned weights initialization (such as Trunc Normal \cite{NEURIPS2019_9015_pytorch} and Kaiming \cite{NEURIPS2019_9015_pytorch} ) in comparison to Vision Transformer (Fig. \ref{fig: effect of weight initialization schemes} in main paper). %the performance improvement on these architectures with our self-supervised weight initialization strategy is less prominent as compared to performance improvement in Vision Transformers.
\begin{table}[!h]
  \begin{minipage}{0.7\textwidth}
   \centering \small
  \setlength{\tabcolsep}{4pt}
  \scalebox{0.8}[0.8]{
  \begin{tabular}{cc|cc}
  \toprule
  \rowcolor{Gray} 
    \textbf{Model}   & \textbf{Initialization} & \textbf{CIFAR100} & \textbf{Tiny-Imagenet}  \\
    \midrule
   ResNet-18 & Trunc Normal \cite{NEURIPS2019_9015_pytorch}  &64.49  & 53.32 \\
   ResNet-18  & Kaiming \cite{NEURIPS2019_9015_pytorch}  &64.08  & 52.19 \\
    % \rowcolor{LightCyan}
   ResNet-18  & Self-supervised (ours)  &\textbf{65.00}  &\textbf{53.48}\\
  \bottomrule
  \end{tabular}}
  \end{minipage}
%   \hfill
  \begin{minipage}{0.28\textwidth}
  %   \vspace{1em}
  \caption{\small Effect of our self-supervised weight initialization scheme on CNNs. }
%   \caption{\small We compare the top-1 accuracy of ResNet-18 initialized with our self-supervision to different initialization schemes. We observe a slight performance gain suggesting that our self-supervised weight initialization scheme is less effective on ConvNets due to the presence of inductive biases. }
 \label{tab: resnet18-self-supervised-weight}
%  \vspace{-1em}
  \end{minipage}

\end{table}
\vspace{2em}

% \section{Visualization of Attention Maps from Self-supervised Learning}
\section{Attention to Salient Input Regions}
\label{sec:appendix-ssl-attn-maps}

\begin{figure}[!h]
 \small \centering
 \begin{minipage}{1.0\textwidth}
 %\fbox{\rule[-.5cm]{0cm}{3cm} \rule[-.5cm]{3cm}{0cm}}
  \includegraphics[width=\linewidth]{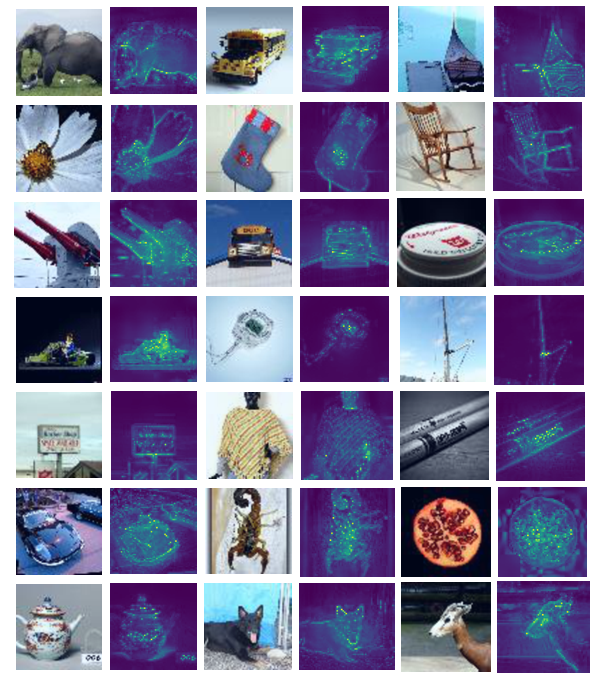}
  \vspace{0.2em}
    \caption{\small \textbf{Self-Supervised Attention:} We visualize the attention of the CLS token from the heads of the last block of ViT on the low-resolution test samples from Tiny-ImageNet. The attention maps show that our self-supervised pre-training learns to segment the class-specific features from unseen test samples without any supervision.
  }
\label{fig: ssl-attn-maps}
  \end{minipage}

\end{figure}

\newpage

% \section{Further comparisons}
\begin{figure}[!h]
 \small \centering
 \begin{minipage}{0.9\textwidth}
 %\fbox{\rule[-.5cm]{0cm}{3cm} \rule[-.5cm]{3cm}{0cm}}
  \includegraphics[width=\linewidth]{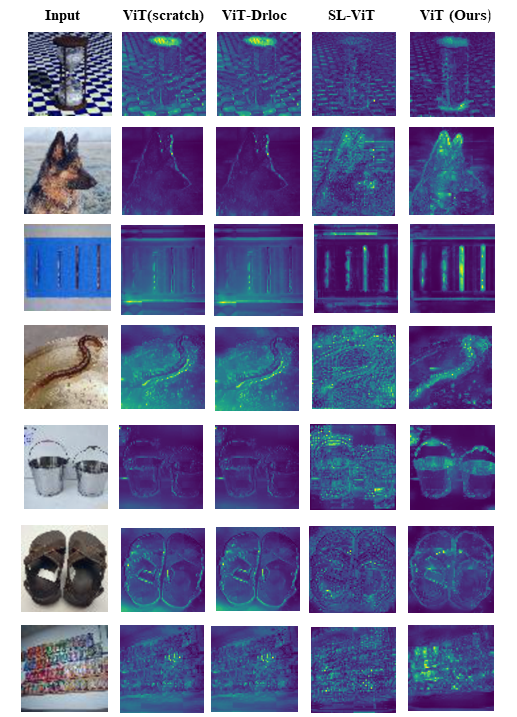}
  \vspace{0.2em}
    \caption{\small \textbf{Supervised Attention:} We visualize the attention of the CLS token from the heads of the last block of ViT across different approaches. All the models are fine-tuned for 100 epochs on Tiny-Imagenet. The attention maps show that our proposed approach is able to capture the salient properties of the specific class in the input and has a sharp focused attention which is missing in other baselines.
  }
\label{fig: finetune-attn-maps}
  \end{minipage}

\end{figure}

\end{document}